\documentclass[11pt,final,3p,sort&compress]{elsarticle}
\usepackage{amsmath}
\usepackage{wasysym}
\usepackage{algorithm}
\usepackage{lmodern}
\usepackage{graphics}
\usepackage{graphicx}
\usepackage{subcaption}
\usepackage{booktabs}
\usepackage{multirow}
\usepackage{array}
\usepackage{textgreek}

\usepackage{amsmath}
\usepackage{soul}
\usepackage{amsthm}
\usepackage{verbatim}
\usepackage{relsize}
\usepackage[dvipsnames]{xcolor}
\usepackage{array}
\usepackage{bm}
\usepackage{caption}
\usepackage{algorithm}
\usepackage{algpseudocode}
\usepackage{enumitem}

\usepackage{hyperref}
\newcommand{\doi}[1]{\textsc{doi}: \href{http://dx.doi.org/#1}{\nolinkurl{#1}}}

\definecolor{comment}{rgb}{0.0, 0.5, 0.0}

\hypersetup{
    colorlinks=true,
    linkcolor=blue,
    filecolor=magenta,
    urlcolor=cyan,
}

\usepackage{booktabs}
\usepackage{xcolor}
\usepackage{color}

\definecolor{mygreen}{rgb}{0,0.6,0}
\definecolor{mygray}{rgb}{0.5,0.5,0.5}
\definecolor{mymauve}{rgb}{0.58,0,0.82}

\usepackage{pdfcomment}

\usepackage{array}
\usepackage{mathtools}
\usepackage{verbatim}
\usepackage{relsize}
\usepackage{multirow}
\usepackage{multicol}

\usepackage{listings}
\usepackage[bitstream-charter]{mathdesign}
\usepackage[T1]{fontenc}
\usepackage[utf8]{inputenc} 

\usepackage[english]{babel}
\usepackage[normalem]{ulem}

\DeclareMathAlphabet{\pazocal}{OMS}{zplm}{m}{n}

\def \balpha {{\pmb{\alpha}}}
\def \bbeta  {{\pmb{\beta}}}

\def \PC {PC$^2$}
\def \bPsi {{\boldsymbol\Psi}}
\def \xxi  {{\boldsymbol{\xi}}}

\def \X  {\pmb{X}} 

\journal{Physics}

\begin{document}

\begin{frontmatter}

\title{Physics-informed Polynomial Chaos Expansion with Enhanced Constrained Optimization Solver and D-optimal Sampling}

\author{Qitian Lu}         
\address{Brno University of Technology, Brno, Czech Republic}
\author{Himanshu Sharma}      
\author{Michael D. Shields}                   
\address{Johns Hopkins University, Baltimore, USA}
\author{Luk{\'a}{\v s} Nov{\'a}k\texorpdfstring{\corref{cor1}}{*}}  \ead{novak.l@fce.vutbr.cz}  \cortext[cor1]{Corresponding author}
\address{Brno University of Technology, Brno, Czech Republic}

\begin{abstract}
Physics-informed polynomial chaos expansions (PC$^2$) provide an efficient physically constrained surrogate modeling framework by embedding governing equations and other physical constraints into the standard data-driven polynomial chaos expansions (PCE) and solving via the Karush-Kuhn-Tucker (KKT) conditions. This approach improves the physical interpretability of surrogate models while achieving high computational efficiency and accuracy. However, the performance and efficiency of PC$^2$ can still be degraded with high-dimensional parameter spaces, limited data availability, or unrepresentative training data. To address this problem, this study explores two complementary enhancements to the PC$^2$ framework. First, a numerically efficient constrained optimization solver, straightforward updating of Lagrange multipliers (SULM), is adopted as an alternative to the conventional KKT solver. The SULM method significantly reduces computational cost when solving physically constrained problems with high-dimensionality and derivative boundary conditions that require a large number of virtual points. Second, a D-optimal sampling strategy is utilized to select informative virtual points to improve the stability and achieve the balance of accuracy and efficiency of the PC$^2$. The proposed methods are integrated into the PC$^2$ framework and evaluated through numerical examples of representative physical systems governed by ordinary or partial differential equations. The results demonstrate that the enhanced PC$^2$ has better comprehensive capability than standard PC$^2$, and is well-suited for high-dimensional uncertainty quantification tasks.

\end{abstract}

\begin{keyword}

Polynomial chaos expansion \sep Physical constraints \sep Surrogate modeling \sep Statistical sampling \sep Lagrange multipliers

\end{keyword}

\end{frontmatter}

\section{Introduction}
\label{sec1}

In many engineering problems, the underlying physical models can be described as ordinary differential equations (ODE) or partial differential equations (PDE) that involve constraints. PDE parameters representing physical characteristics (e.g. material strength) are, in their nature, uncertain. Direct numerical simulation of these models is generally complex and computationally demanding, especially for high-dimensional systems with multiple deterministic and uncertain variables. To alleviate this burden, surrogate models are developed as an efficient approximation of numerical solutions~\cite{Alizadeh2020}. However, constructing data-driven surrogate models often requires a large number of training data to capture the characteristics of the parameter space, which can remain numerically prohibitive or experimentally impractical. In addition, the accuracy of such models is largely determined by the quality of the training data. To reduce dependence on data, some surrogate models incorporate governing equations and physical constraints directly into the training process. These approaches, referred to as physics-informed surrogate models or scientific machine learning (SciML) \cite{karniadakis2021physicsinformed}, leverage prior physical knowledge to improve approximation accuracy and generalization with limited data. 

Recent advances have led to the emergence of representative SciML, such as physics-informed neural networks (PINNs)~\cite{RAISSI2019686}, physics-informed Gaussian processes (PIGPs)~\cite{RAISSI2017683} and physics-informed polynomial chaos expansion (PC$^2$)~\cite{NOVAK2024112926, SHARMA2024117314}. Although all these methods aim to incorporate physical knowledge into surrogate modeling, their implementation strategies differ in important ways. Specifically, PINNs achieve this by adding PDE residual terms to the loss function and minimizing them during training \cite{RAISSI2019686} or defining the specific architecture to obey the constraints \cite{gaynutdinova2025homogenizationguaranteedboundsprimaldual}; PIGPs embed physical operators into the Gaussian process kernel, ensuring that physical constraints are satisfied at the prior level; and PC$^2$ imposes physical constraints directly within the polynomial basis matrix and its derivative matrix, from which the corresponding coefficients can be obtained efficiently by solving the Karush-Kuhn-Tucker (KKT) system~\cite{NOVAK2024112926}. Despite their differences, these methods have achieved considerable success in various physics-informed machine learning tasks~\cite{NOVAK2024112926, SHARMA2024117314, Cai2021, Misyris9282004, Pang2019, TARTAKOVSKY2023967}. However, challenges remain in terms of computational efficiency and performance, particularly when dealing with high-dimensional parameter spaces, limited data availability, or unrepresentative training data. This highlights the need for more efficient approximation solvers, more accurate sampling strategies, and other advanced techniques to further improve the capability of existing SciML. 

Most of the current improvements have focused on PINNs. For accuracy, many efforts have been devoted to developing neural network architectures for specific types of PDEs, e.g. fractional advection-diffusion equations~\cite{Pang2019} or complex geometries \cite{SAHLICOSTABAL2024107324}. The adoption of domain decomposition techniques is proven to be highly effective for complex-coupled problems \cite{li2022gradient} and large-scale computational domains \cite{moseley2023finite}. Further, numerous studies have focused on accelerating PINNs through loss function designs \cite{XIANG202211} or adaptive data sampling ~\cite{guo2022novel, nabian2021efficient}. However, PINNs lack built-in UQ capabilities and are thus not suitable for UQ of physical systems described by parametric PDEs. 

In the context of UQ, the introduction of neural operator further improves the efficiency of surrogate models by directly learning mapping relationships, allowing rapid evaluation across many stochastic realizations without retraining~\cite{Goswami2023}. Compared to the significant progress in neural networks, the advancements in PIGPs and PC$^2$ are relatively limited. Recently, a polynomial chaos expansion (PCE)-based operator learning (OL) framework has been proposed~\cite{sharma2025polynomial}, which extends traditional PCE and PC$^2$ into a new OL paradigm to solve PDE and perform UQ with light computation.

In this paper, motivated by various improvements of PINNs, we focus on adapting the constrained optimization solver and the role of sampling strategies of virtual samples in the regression-based \PC\ framework. First, a straightforward updating of Lagrange multipliers (SULM) solver is proposed to replace the conventional Lagrange multipliers solver, hereafter referred to as the KKT solver. Compared to the KKT solver, the proposed SULM solver avoids constructing and factorizing the full augmented matrix. Instead, it performs sequential updates of the multipliers and coefficients through block elimination, which not only reduces the computational cost but also improves the numerical conditioning. Second, a D-optimal sampling scheme is adopted, using singular value decomposition and QR factorization with column pivoting to select more informative and numerically stable virtual points for surrogate model training. These two techniques are integrated into the PC$^2$ and can be used both individually and in combination, resulting in four configurations: KKT, KKT with D-optimal sampling (KKT-D), SULM, and SULM with D-optimal sampling (SULM-D). The performance and efficiency of these configurations are discussed through several ODE and PDE approximation and UQ tasks.

\section{Methodology}
\label{sec2}

In this section, we briefly present the formulation of the standard PC$^2$ framework based on the KKT solver, followed by the proposed SULM algorithm as its alternative. After that, the D-optimal sampling strategy is introduced to guide the selection of virtual points for more stable training.

\subsection{Physics-Informed Polynomial Chaos Expansion}
\label{subsec21}
Consider a mathematical model $ Y=u\left(\X\right)$, where $\X=(X_1, X_2,\dots, X_M)$ consists of M independent random input variables. To construct a PCE, the random input variables $\X$  are first transformed into a standardized germ $\xxi=(\xi_1, \xi_2,...,\xi_M)$ in order to use polynomial basis according to Wiener-Askey scheme \cite{Askey}. The corresponding multivariate polynomial basis function $ \Psi (\xxi)$ is given by a tensor product of univariate orthonormal polynomials:
\vspace{0.6em}
\begin{eqnarray}
\label{Eq:MultVarPol}
    \Psi_{\balpha} \left( \xxi \right)
    =
    \prod_{i=1}^{M}  \psi_{\alpha_i} \left(\xi_i\right),
\end{eqnarray}
where $ {\balpha}\in \mathbb{N}^M $ represents the multi-index that indicates the polynomial degree associated with each germ dimension $\xi_i$.  Accordingly, the model response $Y$ can be further rewritten as~\cite{ghanem2003stochastic}
\vspace{0.6em}
\begin{eqnarray}
\label{PCE}
    Y  = u\left(\X\right) =
    \sum_{\balpha \in \mathbb{N}^M }
    \beta_{\balpha}\Psi_{\balpha}\left( \xxi\right),
\end{eqnarray}
where $\beta_{\balpha}$ are the deterministic coefficients of multivariate polynomial basis, obtained by minimizing the approximation error using regression techniques such as KKT and SULM solvers or other numerical optimization \cite{SHARMA2024117314}. Note that the basis functions could also be constructed numerically, though it might be challenging to get also derivatives for \PC \cite{OLADYSHKIN2012179,JAKEMANDependentPCE}. 

In a purely data-driven regression-based PCE, the deterministic coefficients $\bbeta$ can be simply estimated using ordinary least squares (OLS) based on the experimental design in the training set. The estimation can be expressed as
\vspace{0.6em}
\begin{eqnarray}
    \bbeta
    =
    \left(\bPsi^{T}\bPsi\right)^{-1} \ \bPsi^{T}  Y,
\label{Eq:PCe_OLS}
\end{eqnarray}
where $Y$ denotes the quantity of interest (QoI) in the training set, corresponding to the model response, and $ \bPsi $ is the data matrix constructed by evaluating the polynomial basis functions at the input samples of the training set.

When additional physical constraints are available, the standard OLS can be extended to incorporate physical information as described in the recent paper \cite{NOVAK2024112926}. Consider a general PDE with its boundary condition:
\vspace{0.6em}
\begin{eqnarray}
    \begin{aligned}
        & \mathcal{L}(\bm{x},t;u(\bm{x},t)) = f(\bm{x},t), & \forall \bm{x}\in \mathcal{D}, t \in \mathcal{T}\\
        & \mathcal{B}(\bm{x},t;u(\bm{x},t)) = g(\bm{x},t), & \forall \bm{x}\in \partial\mathcal{D}, t \in \mathcal{T}
        \label{eqn:PDE UQ}
    \end{aligned}
\end{eqnarray}
where $\mathcal{L} (\cdot)$ is the differential operator defined in the spatial domain $\mathcal{D}$ and the temporal domain $\mathcal{T}$, with the PDE source term $f$; $\mathcal{B} (\cdot)$ is the boundary operator defined at the spatial boundary $\partial\mathcal{D}$  and the temporal domain $\mathcal{T}$, with the boundary source term $g$.  Similarly, the initial operator can be defined in the same manner as the boundary operator and is therefore omitted here for brevity. Such a constrained optimization problem can be formulated as
\vspace{0.6em}
\begin{eqnarray}
    \begin{aligned}
        & \mathcal{M} (\bbeta) = \min_{\bbeta} \sum_{j=1}^{n_{\mathrm{sim}}} \left[Y^j - u^{\mathrm{PCE}} \left(\bm{x}^j, t^j \right) \right]^2=  \min_{\bbeta} \lVert Y  - \bPsi\bbeta \rVert^2\\
        \text{s.t. } &  \mathcal{L}\left(\bm{x}_{\mathrm{V}},t_{\mathrm{V}}; u\left(\bm{x}_{\mathrm{V}},t_{\mathrm{V}}\right)\right) = f\left(\bm{x}_{\mathrm{V}},t_{\mathrm{V}}\right), \\
        & \mathcal{B}\left(\bm{x}_{\mathrm{V}},t_{\mathrm{V}}; u\left(\bm{x}_{\mathrm{BC}},t_{\mathrm{BC}}\right)\right) = g\left(\bm{x}_{\mathrm{BC}},t_{\mathrm{BC}}\right).
    \end{aligned}
    \label{Eq:PC2_UQ_definition}
\end{eqnarray}
Here, we have three different types of points, and $n_{\mathrm{sim}}$, $n_{\mathrm{V}}$, and $n_{\mathrm{BC}}$ represent the numbers of data points in the experimental design, virtual points sampled from the PDE equation, and boundary points defined by the boundary conditions, respectively. The estimation of deterministic coefficients $\bbeta$ can then be performed efficiently through constrained least squares using Lagrange multipliers. We first construct the following Lagrangian function:
\vspace{0.6em}
\begin{eqnarray}
 L(\bbeta, \lambda)
= 
\frac{1}{2}\mathcal{M}(\bbeta)
+ \sum_{i=1}^{n_{\mathrm{BC}}} 
\lambda_{i}\left( a_{i}^{T}\bbeta- c_{i_\mathrm{BC}} \right)
+ \sum_{j=1}^{n_{\mathrm{V}}} 
\lambda_{j}\left( a_{j}^{T}\bbeta - c_{j_\mathrm{v}} \right),
\label{Eq:Lagrangian function}
\end{eqnarray}
where ${\lambda_i}$ and ${\lambda_j}$ are the Lagrange multipliers, ${a_i}$ and ${a_j}$ are calculated by the polynomial basis or the derivative of polynomial basis. The boundary conditions, restricted by the boundary operator $\mathcal{B}$, are evaluated at the $n_{\mathrm{BC}}$ boundary points $\xxi_{\mathrm{BC}}$. The corresponding reference vector $\mathbf{c}_{\mathrm{BC}}=(c_1,c_2,\dots,c_{n_\mathrm{BC}} )$ is determined by the boundary source term $g$. The differential equation constraint, represented by the differential operator $\mathcal{L}$, is evaluated at the $n_{\mathrm{V}}$ virtual points $\xxi_{\mathrm{V}}$. The corresponding reference vector $\mathbf{c}_{\mathrm{V}}=(c_1,c_2,\dots,c_{n_\mathrm{v}} )$ is determined by the PDE source term $f$. After assembling all the prescribed constraints into a matrix $\mathbf{A}$, the following KKT system of linear equations can be obtained, which extends the form of the standard OLS:
\vspace{0.6em}
    \begin{eqnarray}
        \underbracket[0.4pt]{\begin{bmatrix}
        \boldsymbol\Psi^{T}\boldsymbol\Psi & \mathbf{A}^{T} \\
        \mathbf{A} & \boldsymbol{0}  
        \end{bmatrix}}_{\mathrm{KKT \; matrix }}
        \begin{bmatrix}
        \bbeta  \\
        \boldsymbol\lambda
        \end{bmatrix}
        =
        \begin{bmatrix}
        \boldsymbol\Psi^{T}  Y  \\
        \mathbf{c}
        \end{bmatrix}.
        \label{Eq:KKT_system}
    \end{eqnarray}
    
\subsection{Straightforward Updating of Lagrange Multipliers}
\label{subsec22}
In high-dimensional parametric problems, directly solving the KKT system shown in Eq. (\ref{Eq:KKT_system}) becomes computationally expensive. On the one hand, the minimum number of polynomial basis required grows rapidly with increasing dimensionality. On the other hand, the complexity of high-dimensional systems typically demands a larger number of experimental design points and physics-constrained points to ensure convergence. To overcome this limitation, a reformulated solver based on the straightforward updating of Lagrange multipliers (SULM) is introduced, as proposed in~\cite{SCOTT2022}. Instead of assembling and factorizing the KKT system that contains the full information, the SULM solver decomposes the optimization process into a sequence of lightweight matrix operations. This restructuring not only reduces computational cost but also mitigates potential ill-conditioning hidden in large KKT systems.

The SULM solver begins by solving the unconstrained least squares problem, which provides an initial estimation of the deterministic coefficients without considering any physical constraint. This preliminary coefficient vector, denoted as $\tilde{\boldsymbol{\beta}}$, follows the same formulation as Eq. (\ref{Eq:PCe_OLS})): 
\vspace{0.6em}
\begin{equation}
\tilde{\boldsymbol{\beta}}
= \left( \boldsymbol{\Psi}^{T}\boldsymbol{\Psi} \right)^{-1} \boldsymbol{\Psi}^{T}\mathbf{Y}.
\label{eq:unconstrained_ls}
\end{equation}
This unconstrained solution serves as the baseline for enforcing the physical constraints represented by the matrix $ \mathbf{A}$. Therefore, an updating operator is introduced to adjust the preliminary coefficients $\tilde{\boldsymbol{\beta}}$ toward a physically consistent solution:
\vspace{0.6em}
\begin{equation}
\mathbf{J} = -\,\left( \boldsymbol{\Psi}^{T}\boldsymbol{\Psi} \right)^{-1}\mathbf{A}^{T}.
\label{eq:J_matrix}
\end{equation}
Based on this operator, the reduced constraint matrix can be established to describe the interaction between the constraints. It is defined as
\vspace{0.6em}
\begin{equation}
\mathbf{Y}_{\mathrm{c}} = \mathbf{A}\mathbf{J}
\label{eq:reduced constraint matrix}
\end{equation}
To evaluate the deviation of $\tilde{\boldsymbol{\beta}}$ from the prescribed constraints, a residual vector is then calculated by
\vspace{0.6em}
\begin{equation}
\mathbf{r} = \mathbf{c} - \mathbf{A}\tilde{\boldsymbol{\beta}}.
\label{eq:residual vector}
\end{equation}
With the reduced constraint matrix $\mathbf{Y}_{\mathrm{c}}$ and residual vector $\mathbf{r}$, the Lagrange multipliers can be efficiently obtained by solving the following linear system:
\vspace{0.6em}
\begin{equation}
\mathbf{Y}_{\mathrm{c}}\boldsymbol\lambda = \mathbf{r}.
\label{eq:lambda}
\end{equation}
Finally, the deterministic coefficients that satisfy all physical constraints are updated using the Lagrange multipliers through
\vspace{0.6em}
\begin{equation}
\boldsymbol{\beta} = \tilde{\boldsymbol{\beta}} + \mathbf{J}\boldsymbol\lambda.
\label{eq:beta_update}
\end{equation}

Specifically, if we have $n_{\mathrm{sim}}$ samples, $p$ polynomial bases, and $n_{\mathrm{c}}$ constraints, i.e. $\boldsymbol{\Psi} \in \mathbb{R}^{n_{\mathrm{sim}} \times p}$ and $\mathbf{A} \in \mathbb{R}^{n_{\mathrm{c}} \times p}$, the theoretical computational cost of the KKT solver is $O(n_{\mathrm{sim}}p^{2}+(p{+}n_{\mathrm{c}})^{3})$, while that of the SULM solver is $O(n_{\mathrm{sim}}p^{2}+p^{3} + p^{2}n_{\mathrm{c}} + pn_{\mathrm{c}}^{2} + n_{\mathrm{c}}^{3})$. The time advantage of the SULM solver arises from its decoupling of the constraints from the full KKT system. Moreover, this decoupled structure also allows possible exploitation in active learning frameworks, as Eq.~(\ref{eq:unconstrained_ls}) can be evaluated only once and reused for retraining.

\subsection{D-optimal Sampling}
\label{subsec23}
Regardless of whether a surrogate model is purely data-driven or physics-informed, the quality of the training samples directly influences the accuracy and stability of the results. When sufficient data are available, this quality is highly determined by the sampling strategy adopted during their generation. The crude Monte Carlo Sampling (MC) is one of the earliest and most general sampling approaches. It is convenient for application but computationally expensive especially in high-dimensional spaces, demanding a large number of samples to cover the parameter space. To enhance the sampling efficiency and accuracy, much effort has been devoted to generating more uniformly distributed samples with lower discrepancy, such as direct discrepancy optimization and Quasi Monte Carlo Sampling (QMC) method~\cite{vovrechovsky2020modification, fang1993number, hickernell1998lattice}. Another idea of sampling optimization focuses on sampling variance reduction, where stratified sampling techniques, such as Latin Hypercube Sampling (LHS)~\cite{McKay01051979}, are commonly used. In addition to improving statistical representativeness, space-filling criteria have been developed to enhance the geometric uniformity of sampling space, which are based on the distance of the sampling points. These criteria, such as Maximin and miniMax criteria~\cite{johnson1990minimax}, phi-criteria~\cite{morris1995exploratory}, and periodic distance-based criteria~\cite{vovrechovsky2020modification}, effectively avoid both clustering of points and large empty regions within sampling space. In comparison, D-optimal sampling focuses on maximizing the information of the sample matrix, thereby leading to better numerical conditioning and stable regression performance.

Similarly to PINNs~\cite{WU2023115671}, sampling strategies are also essential in PC$^2$ to ensure numerical stability and accuracy. In the scenarios of limited data, the selection of virtual points becomes particularly critical, as these points are sampled to simulate the PDE and thus represent the most necessary physical information of the system. A well-chosen virtual point set should capture the dominant characteristics of the parametric space while minimizing redundancy. To achieve this, a D-optimal strategy is employed to maximize the information density of the selected virtual points.

To construct a D-optimal subset of virtual points, a large pool of candidate samples is first generated to provide sufficient coverage of the parametric domain. The number of candidate points is typically several times larger than the number required for surrogate modeling~\cite{loeppky2009choosing}. In this study, we generate a candidate matrix of virtual points $\Psi_{V}$ by simple random sampling of $kn_{\mathrm{V}}$ virtual points, where the oversampling ratio $k$  (here, we set $k=3$) controls the size of the candidate pool. After that, an algorithm based on singular value decomposition (SVD) and QR factorization with column pivoting is adopted to identify the optimal subset from this candidate pool. The SVD of the transposed candidate matrix can be formulated as 
    \vspace{0.6em}
    \begin{eqnarray}
    {\Psi_{V}}^T = \mathbf{U} \mathbf{\Sigma} \mathbf{V}^T
    \label{Eq: SVD}
    \end{eqnarray}
where $\mathbf{U}$ and $\mathbf{V}$ are orthogonal matrices, referred to as left and right singular vector matrices, respectively, and $\mathbf{\Sigma}$ is a diagonal matrix containing singular values. To extract an informative subset of virtual points, the right singular vector $\mathbf{V}$ is ordered using QR factorization with column pivoting.
   \vspace{0.6em}
    \begin{eqnarray}
    \mathbf{VP} = \mathbf{Q} \mathbf{R}
    \label{Eq: pivoting}
    \end{eqnarray}
where $\mathbf{P}$ is the column permutation matrix, $\mathbf{Q}$ is the orthogonal matrix, and $\mathbf{R}$ is the upper triangular matrix. The pivoting step ranks the candidate points by the column norms of the right singular vectors, selecting those that provide the greatest numerical independence for D-optimal sampling.

Finally, the first $n_{\mathrm{V}}$ rows of the pivoting matrix $\mathbf{P}$, corresponding to the most informative virtual points, are selected and used in the training of the surrogate model.

\section{Numerical Results}
\label{sec3}

In this section, we compare the performance of two constrained optimization solvers: the original Lagrange multipliers (i.e., KKT) and SULM across both deterministic and stochastic tasks, including the cases of increasing dimensionality. In addition, we also investigate the impact of D-optimal sampling of virtual points on the accuracy and efficiency of surrogate modeling. All compared methods are extended by $p$ adaptivity governed by $\epsilon$ consisting of mean squared error in the approximated function (``data error''), error associated with failure to obey the PDE constraints (``PDE error''), and error in the boundary conditions (``BC error'') \cite{NOVAK2024112926}. For plotting, the absolute error (AE) and relative absolute error (RAE) are adopted to evaluate the error distribution in the field, whereas the mean squared error (MSE) is used as a global indicator of model accuracy:
\vspace{0.6em}
\begin{eqnarray}
    \begin{aligned}
        &\mathrm{AE}=\lvert y_{pce}-y_{ref}\rvert,\quad\mathrm{RAE}=\frac{\lvert y_{pce} - y_{ref}\rvert }{\lvert y_{ref} \rvert}, \\
        &\quad\quad\quad\mathrm{MSE} = \frac{1}{N} \sum_{i=1}^{N} (y_{pce} - y_{ref})^2.
    \end{aligned}
    \label{Eq. error}
\end{eqnarray}
To eliminate the computational cost of insignificant polynomial bases, hyperbolic truncation \cite{BLATMAN20112345} strategies are employed in those cases with Karhunen Loève (KL) expansion. Numerical results of KKT algorithm are obtained directly from the original implementation of PC$^2$ in UQPy \cite{TSAPETIS2023101561}. 

\subsection{Toy Example: 1D Euler Equation with Random Constant Load}
\label{subsec31}
The first toy example aims to verify the equivalence between the KKT and SULM algorithms in a relatively simple scenario: the approximation of 1D Euler equation, which involves a deterministic input variable (the spatial coordinate $x$) and a random constant load $q$ following a uniform distribution:
\vspace{0.6em}
\begin{eqnarray}
    \begin{aligned}
        &\frac{\partial^4 u(x,q)}{\partial x^4} + q = 0, \quad x \in [0,1],~~q \sim \mathcal{U}[1,2], \\
        &u(0,q) = u(1,q) = u_{xx}(0,q) = u_{xx}(1,q) = 0.
    \end{aligned}
    \label{Eq. 1D Euler Equation}
\end{eqnarray}
Since this example is straightforward and its analytical solution can be represented exactly in polynomial form, we do not introduce any additional training data in surrogate modeling. Instead, virtual points are provided to ensure sufficient information for convergence, which can also be regarded as scenarios of extremely scarse or expensive training data.

\begin{figure}[t!]
    \centering
    \includegraphics[width=0.5\textwidth]{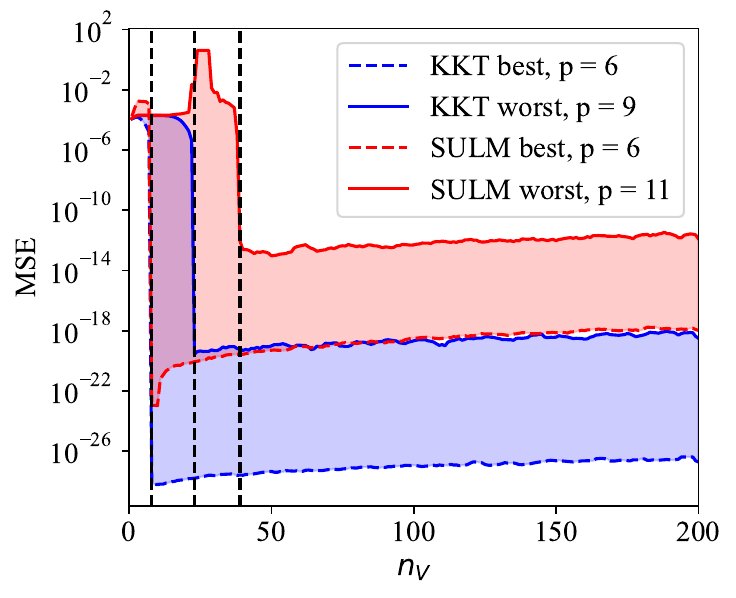}
    \caption{Comparison of KKT and SULM solvers for the 1D Euler equation with a random constant load.  The vertical dashed lines mark the critical value of $n_{\mathrm{V}}$ separating the underdetermined and overdetermined regimes of the surrogate model.}
    \label{fig:EB_RL}
\end{figure}

For PC$^2$, we investigate the polynomial order $p$ ranging from 6 to 11 with the number of boundary points fixed at $n_{\mathrm{BC}}=40$. As the virtual points $n_{\mathrm{V}}$ increase from 1 to 200, the MSE corresponding to the KKT and SULM solvers, each with and without D-optimal sampling, is calculated for comparison. In this example, the influence of D-optimal sampling on computational cost and accuracy is almost negligible, as the problem can be well approximated with a few virtual points, thereby D-optimal sampling is not applied in this case.  However, the results of two constrained optimization solvers still have some differences. As shown in Fig. \ref{fig:EB_RL}, although the MSE magnitudes of both solvers are extremely small (lower than $10^{-12}$ after convergence), it can be observed that the MSE of SULM is obviously higher compared to KKT, which is caused by the accumulation of machine errors due to additional matrix operations. Since the level of target accuracy in engineering applications is significantly higher, this limitation of SULM doesn't represent a significant drawback. Interestingly for both solvers, the best performance is achieved at $p=6$, while the worst results are obtained at $p=9$ and $p=11$ for KKT and SULM, respectively. Moreover, all solvers can converge in just a few seconds, so it is not meaningful to discuss the impact of solvers on computational cost here. Therefore, it can be concluded that, in simple examples, SULM algorithm leads to equivalent performance in comparison to KKT.

\subsection{2D Heat Equation: Dirichlet BC and Random Thermal Diffusivity Coefficient}
\label{subsec32}

The second example is designed to show the difference between SUML and KKT with respect to computational efficiency. In this example, we discuss the following 2D heat equation with Dirichlet boundary and uniformly distributed coefficient of thermal diffusivity $\mathcal{D}$:
\vspace{0.6em}
\begin{equation}
    \begin{aligned}
        \frac{\partial u(x,y, t)}{\partial t}
        &=\mathcal{D} \left( \frac{\partial^2u(x,y,t)}{\partial x^2}
        +\frac{\partial^2u(x,y,t)}{\partial y^2} \right), 
        \qquad x,y \in [0, 1], \ t \in [0, 1], \ \mathcal{D}\sim    \mathcal{U}[0.001,0.1]\\
         u(0,y,t) &= u(1,y,t)= u(x,0,t)= u(x,1,t)=0, \quad u(x,0) =  \sin (2\pi x)\sin(2\pi y),
    \end{aligned}
    \label{Eq:2D_HE_DB}
\end{equation}

\begin{figure}[htbp]
    \centering
    \begin{subfigure}{0.5\textwidth}
        \centering        \includegraphics[width=\linewidth]{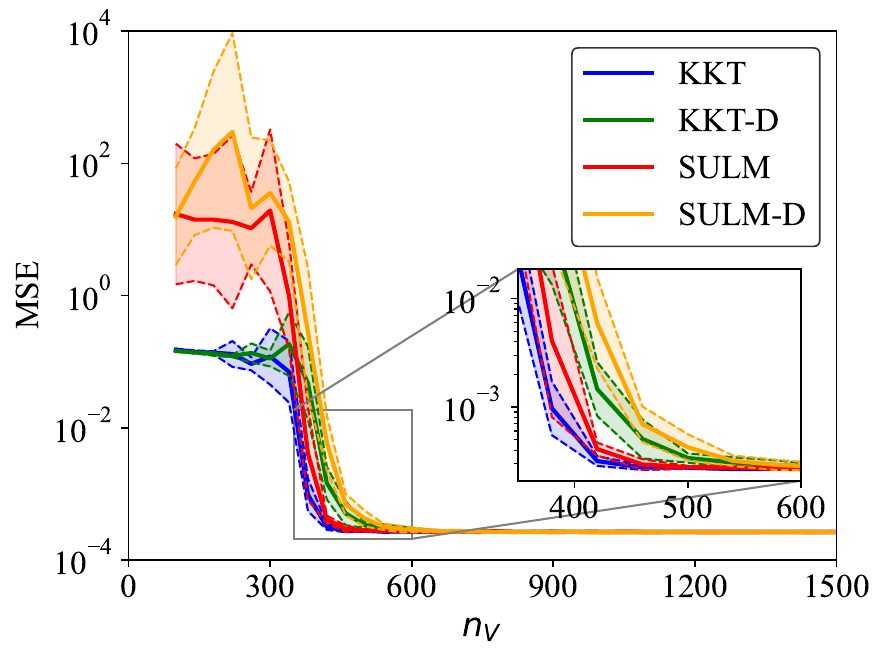}
        \caption{}
        \label{fig:HEDB_MSE}
    \end{subfigure}
    \begin{subfigure}{0.47\textwidth}
        \centering
        \includegraphics[width=\linewidth]{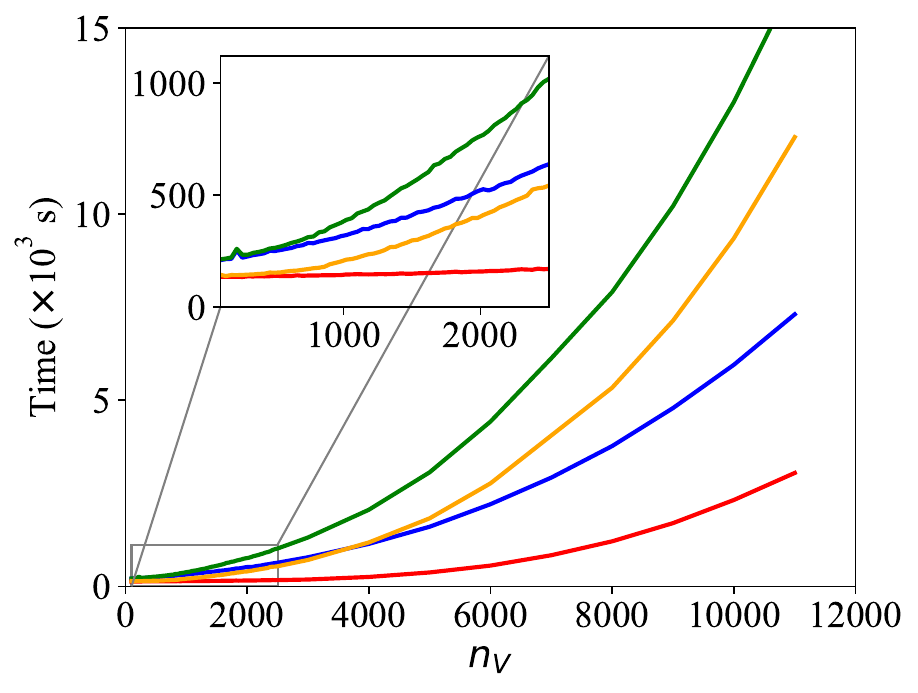}
        \caption{}
        \label{fig:Time_HEDB}
    \end{subfigure}
    \caption{Comparison of (a) computational accuracy and (b) computational cost of KKT and SULM solvers, with and without D-optimal sampling, for 2D heat equation with Dirichlet boundary.}
\end{figure}

For PC$^2$, we set $p=12$, $n_{\mathrm{BC}}=1000$, and introduce $n_{init}=1000$ as the number of initial points, with $n_{\mathrm{V}}$ gradually increasing from 100. For each $n_{\mathrm{V}}$, we perform 10 independent runs, from which the mean and standard deviation of the MSE are calculated. As shown in Fig.~\ref{fig:HEDB_MSE}, all cases exhibit a rapid convergence, approaching an MSE of the order $10^{-4}$. When $n_{\mathrm{V}}$ is small, the MSE of SULM and SULM-D are considerably higher than those of KKT and KKT-D. Nevertheless, KKT and SULM, as well as KKT-D and SULM-D, almost converge at the same $n_{\mathrm{V}}$, although the introduction of D-optimal sampling leads to a slight delay in convergence. Ultimately, all four strategies reach the same error level. Fig.~\ref{fig:Time_HEDB} illustrates the growth of computational cost with increasing $n_{\mathrm{V}}$, including a magnified view near the critical convergence region. Since the minimum $n_{\mathrm{V}}$ required for convergence is relatively small, the influence of D-optimal sampling on computational cost is negligible near the critical convergence region; the observed differences are primarily attributed to the KKT and SULM solvers. The ranking of computational efficiency in this regime is SULM, SULM-D, KKT, and KKT-D. However, as $n_{\mathrm{V}}$ increases after convergence, the cost introduced by D-optimal sampling becomes more significant. At $n_{\mathrm{V}}=4000$, the costs of SULM-D and KKT are nearly identical, indicating that the additional cost of D-optimal sampling eliminates the efficiency gap between the KKT and SULM solvers. With further increases in $n_{\mathrm{V}}$, the cost induced by D-optimal sampling becomes increasingly dominant, and the efficiency ranking shifts to and stabilizes as SULM, KKT, SULM-D, and KKT-D. Overall, under the same PCE setting, SULM demonstrates significantly better computational efficiency compared to KKT, while maintaining comparable accuracy, particularly in large-scale computations.

\subsection{2D Heat Equation: Neumann BC and Random Thermal Diffusivity Coefficient}
\label{subsec33}

In order to clearly show the benefits of optimal sampling and its role for each algorithm, we consider the following heat equation with Neumann boundary and random coefficient of thermal diffusivity $\mathcal{D}$:
\vspace{0.6em}
\begin{equation}
\begin{aligned}
    \frac{\partial u(x,y, t)}{\partial t}
    &=\mathcal{D} \left( \frac{\partial^2u(x,y,t)}{\partial x^2}
    +\frac{\partial^2u(x,y,t)}{\partial y^2} \right), 
    \qquad x,y \in [0, 1], \ t \in [0, 1], \ \mathcal{D}\sim \mathcal{U}[0.001,0.1]\\
    u_x(0,y,t) &= u_x(1,y,t)= u_y(x,0,t)= u_y(x,1,t)=0, \quad u(x,0) = 0.5(\sin (4\pi x)+\sin(4\pi y)),
\end{aligned}
\label{Eq:2D_HE_NB}
\end{equation}
Compared to the Dirichlet boundary, the Neumann boundary specifies the value of the partial derivative at the boundary rather than the solution itself. Consequently, it offers weaker control over the training process and may lead to boundary value drift. without introducing additional training data, more virtual points are required to enhance the PDE constraints on the solution in this example. 

\begin{figure}[htbp]
    \centering
    \begin{subfigure}{0.49\textwidth}
        \centering        \includegraphics[width=\linewidth]{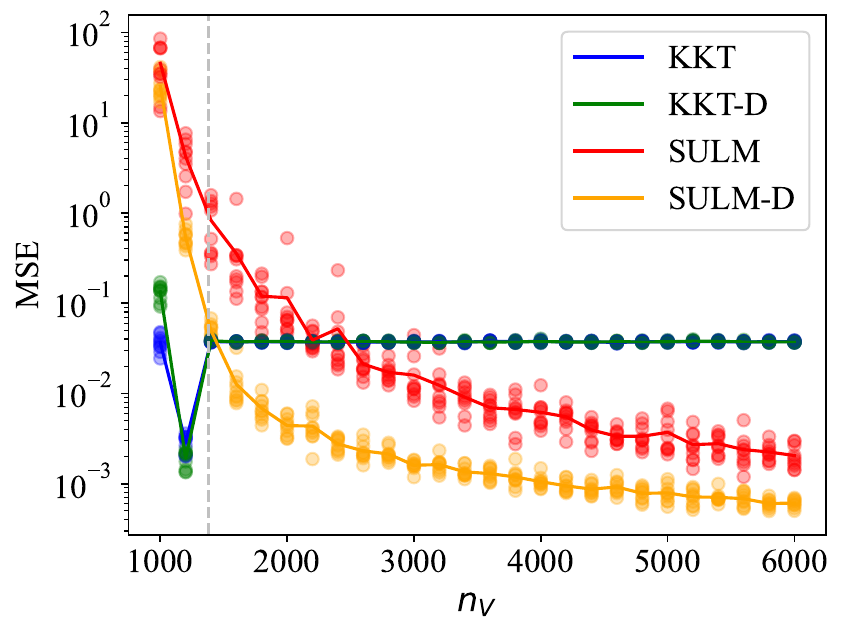}
        \caption{}
        \label{fig:HENB_MSE}
    \end{subfigure}
    \begin{subfigure}{0.49\textwidth}
        \centering
        \includegraphics[width=\linewidth]{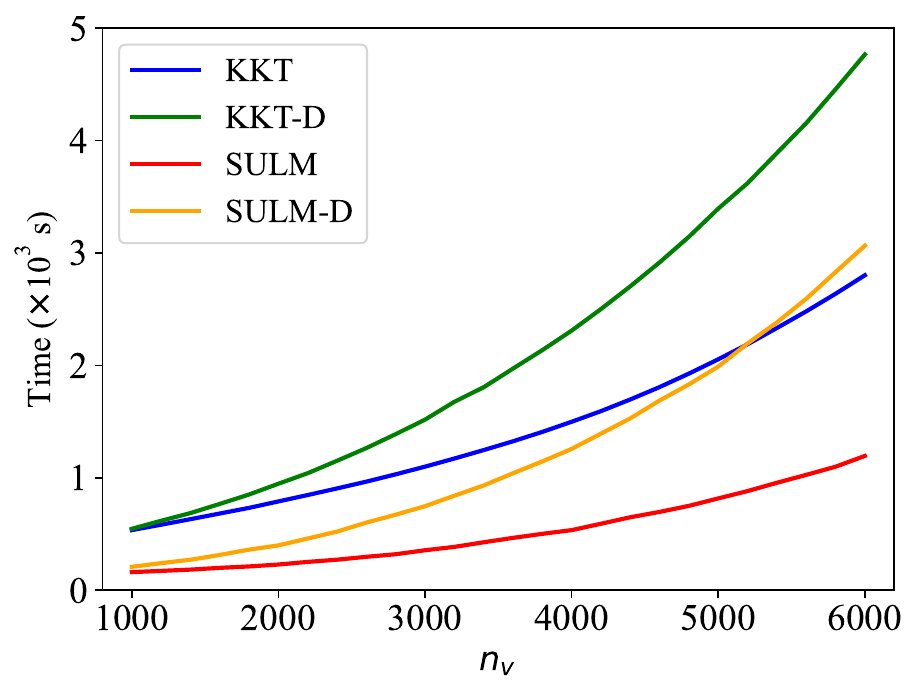}
        \caption{}
        \label{fig:Time_HENB}
    \end{subfigure}
    \caption{Comparison of (a) computational accuracy and (b) computational cost of KKT and SULM solvers, with and without D-optimal sampling, for 2D heat equation with Neumann boundary. The vertical dashed line in (a) indicates the critical value of $n_{\mathrm{V}}$ beyond which the overconstrained surrogate model reaches a stable regime and the error no longer changes.}
\end{figure}

\begin{figure}[t!]
    \centering
    \includegraphics[width=1\textwidth]{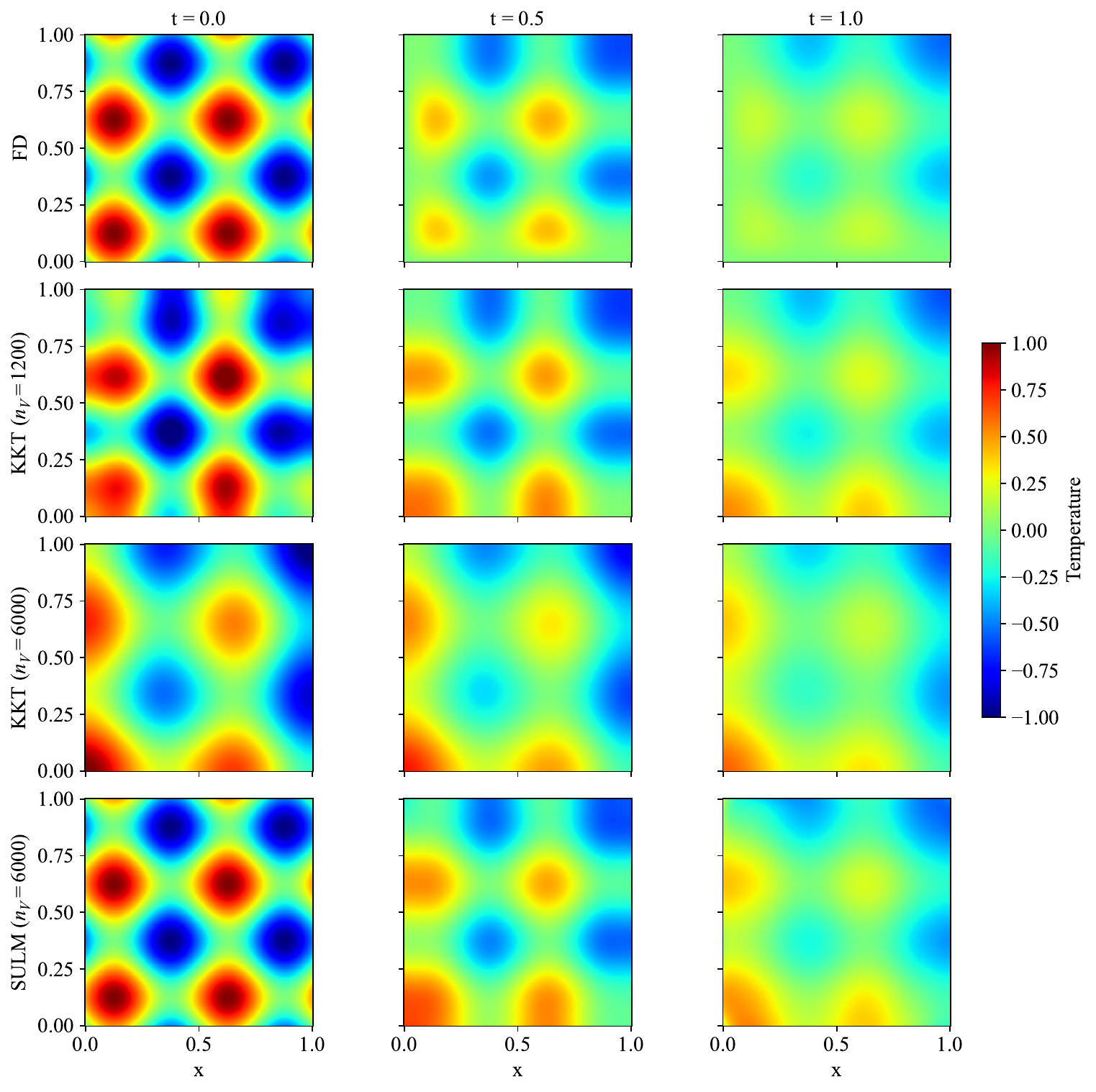}
    \caption{Comparison of the temperature fields solved by FD, KKT ($n_{\mathrm{V}}=1200$ and $n_{\mathrm{V}}=6000$) and SULM ($n_{\mathrm{V}}=6000$).}
    \label{fig:HENB_field}
\end{figure}

\begin{figure}[t!]
    \centering
    \includegraphics[width=1\textwidth]{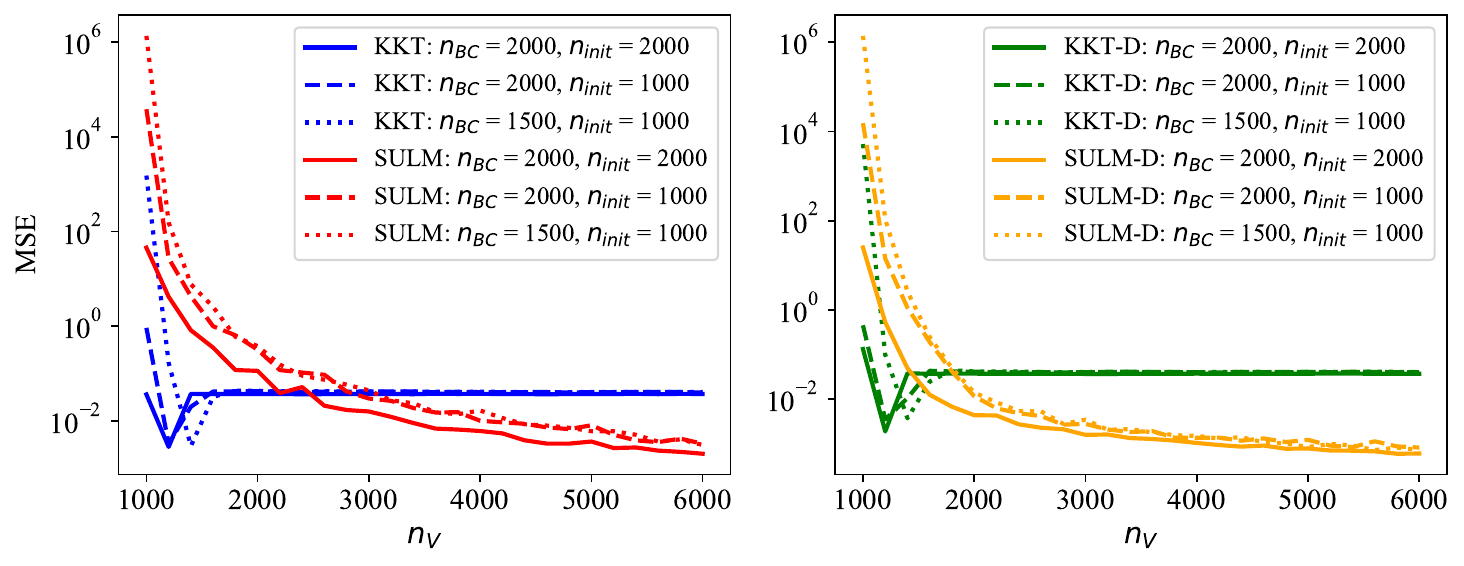}
    \caption{MSE versus $n_{\mathrm{V}}$ for the KKT, KKT-D, SULM and SULM-D solvers for 2D heat equation with Neumann boundary with different $n_{\mathrm{BC}}$ and $n_{\mathrm{init}}$.}
    \label{fig:MSE_diff_bc_init}
\end{figure}

For PC$^2$, we set $p=14$, $n_{\mathrm{BC}}=2000$, and $n_{init}=2000$. We test the case of $n_{\mathrm{V}}$ from 1000 to 6000. For each $n_{\mathrm{V}}$, we perform 10 independent runs, and the distributions of MSE are reported. As shown in Fig.~\ref{fig:HENB_MSE}, similar to the case with Dirichlet boundary, when $n_{\mathrm{V}}$ is small, the MSE of SULM is significantly higher than that of KKT. As $n_{\mathrm{V}}$ increases, the MSE of SULM gradually decreases and eventually approaches an asymptotic error plateau. In comparison, the MSE of KKT rapidly drops to the order of $10^{-3}$ at $n_{\mathrm{V}}=1200$, but subsequently shows a rebound and stabilizes at $10^{-2}$. To further investigate the source of this phenomenon, Fig.~\ref{fig:HENB_field} presents the solution field solved by the implicit Euler finite difference (FD) method, KKT ($n_{\mathrm{V}}=1200$ and $n_{\mathrm{V}}=6000$), and SULM ($n_{\mathrm{V}}=6000$) at $t=0$, $t=0.5$ and $t=1$ for $\mathcal{D}=0.01$. The FD result serves as the reference numerical solution, employing a uniform grid of $200\times200$ nodes in the spatial domain and a time step of $\Delta t=10^{-4}$, which is to ensure the numerical stability. When $t=0$, the SULM result is in perfect agreement with the FD reference, satisfying the initial condition of the PDE. In comparison, the KKT solution ($n_{\mathrm{V}}=1200$) performs slightly worse near the boundaries at $x=0$ and $y=0$. The KKT ($n_{\mathrm{V}}=6000$), which emphasizes enforcing the PDE constraint, exhibits relatively larger errors throughout the domain. At $t=0.5$, both the results of SULM and KKT ($n_{\mathrm{V}}=1200$) start to show small boundary discrepancies against the FD solution, although their overall accuracy remains acceptable. However, in the over-constrained KKT case ($n_{\mathrm{V}}=6000$), the solution exhibits noticeable boundary errors and temporal homogenization, where only negligible evolution can be observed with increasing $t$. By $t=1$, the three strategies give comparable results, and the errors are maintained at a similar level. We also investigate the influence of varying $n_{\mathrm{BC}}$ and  $n_{\mathrm{init}}$ on the over-constrained threshold in KKT and on the MSE behavior of both KKT and SULM. For KKT, $n_{\mathrm{BC}}$ corresponds to first-order derivative under the Neumann boundary and thus serves as a hard constraint, while $n_{\mathrm{init}}$ acts only as a soft constraint. As shown in Fig.~\ref{fig:MSE_diff_bc_init}, reducing $n_{\mathrm{BC}}$ delays the over-constrained threshold without affecting the MSE, while reducing $n_{\mathrm{init}}$ only leads to negligible changes. In contrast, SULM proceeds by iteratively moving from an unconstrained solution toward a constrained one. As a result, it is more sensitive to the value of $n_{\mathrm{init}}$, and reducing $n_{\mathrm{init}}$ leads to a relatively pronounced degradation in MSE. In addition, the computational cost follows the same trend as in the previous case in Fig.~\ref{fig:Time_HENB}. The efficiency of SULM is significantly better than KKT, while the impact of D-optimal sampling on cost is highly dependent on the adopted $n_{\mathrm{V}}$. These findings further verify the computational efficiency of the SULM solver, while highlighting the limitations of the KKT solver in specific scenarios. Moreover, from the errors of individual runs, it can be seen that D-optimality leads to consistent convergence without outliers and thus reduce the dependence of \PC on specific positions of virtual samples. This suggests the potential of replacing the KKT solver by combining the SULM with D-optimal sampling, especially in more complex cases. 

\subsection{1D Euler Equation with KL-Decomposed Stochastic Bending Stiffness}
\label{subsec34}
From this example, we now turn to the application of the two solvers and the D-optimal sampling to the cases with increasing dimensionality. Consider the following 1D Euler equation with random bending stiffness $EI$:
\vspace{0.6em}
\begin{eqnarray}
    \begin{aligned}
        &\frac{\mathrm{d}^2}{\mathrm{~d} x^2}\left(EI(X,\xi_i) \frac{\mathrm{d}^2 u}{\mathrm{~d} x^2}\right)=q,, \quad x \in [0,L],\quad i=,1,2,...,5, \\
        &\quad\quad\quad u(0,q) = u(1,q) = u_{xx}(0,q) = u_{xx}(1,q) = 0.
    \end{aligned}
    \label{Eq. 1D Euler Equation with KL}
\end{eqnarray}
Here, the random field $EI(X,\xi_i)$ is obtained by the Karhunen Loève (KL) expansion, given as
\vspace{0.6em}
\begin{equation}
    EI(\mathbf{x}, \theta)= {\overline {EI}}+\sum_{i=1}^{5} \sqrt{\lambda_{i}} \phi_{i}(\mathbf{x}) \xi_{i}(\theta),
    \label{eq:KL_expansion}
\end{equation}
where $\lambda_i$ and $\phi_i(x)$ denote the eigen values and vectors of the covariance kernel, respectively, and $\xi_i(\theta)$ is an independent standard Gaussian random variable with correlation length $l_c=5\ m$. In this case, the first five eigenmodes are adopted to capture more than 99\% cumulative variance of the random field. The mean and standard deviation of the bending stiffness are $\overline {EI}=80\times10^{-4}\ GPa$ and $\sigma=0.05\overline {EI}$. We set the length of the beam $L=10\ m$ and the distributed load $q=-5\ KN/m$.

\begin{figure}[p]
    \centering
    \begin{subfigure}{0.70\textwidth}
        \centering
        \includegraphics[width=\linewidth]{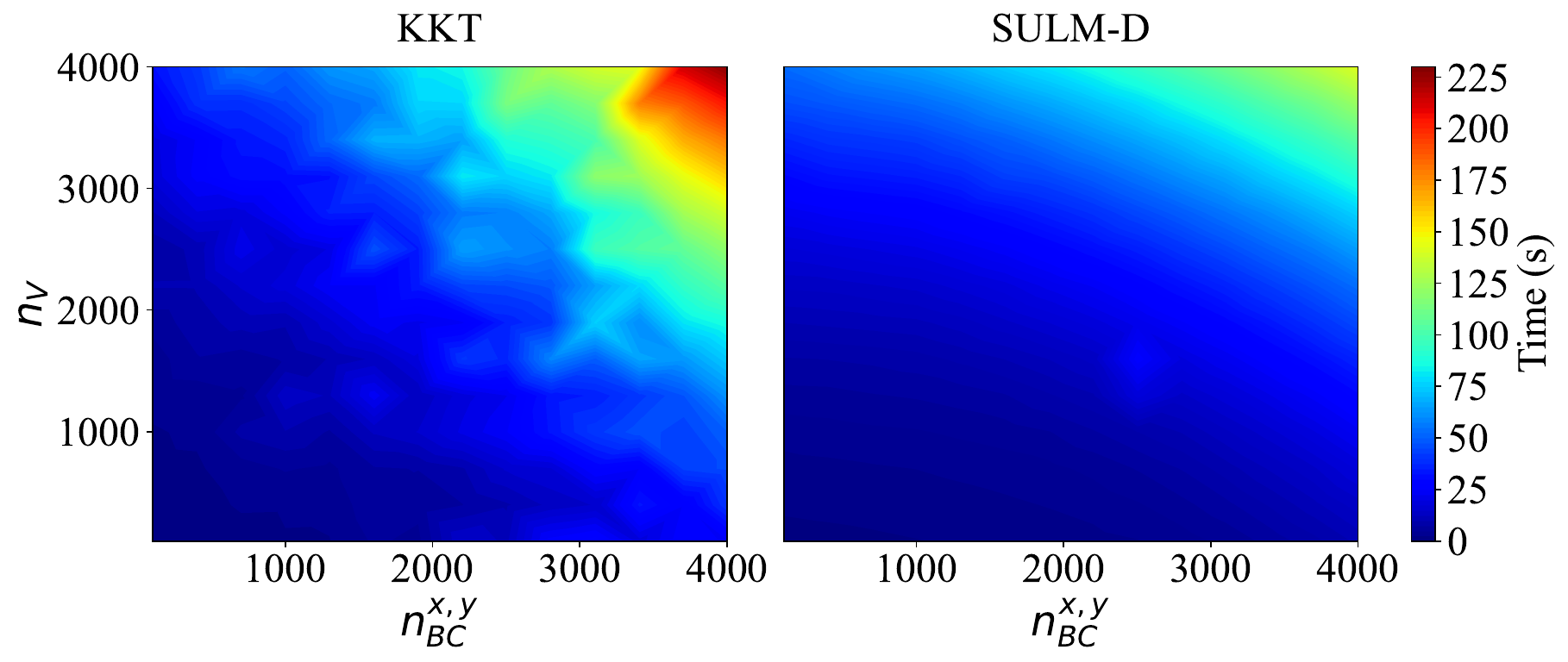}
        \caption{}
        \label{fig:EBKL_two}
    \end{subfigure}
    \begin{subfigure}{0.70\textwidth}
        \centering
        \includegraphics[width=\linewidth]{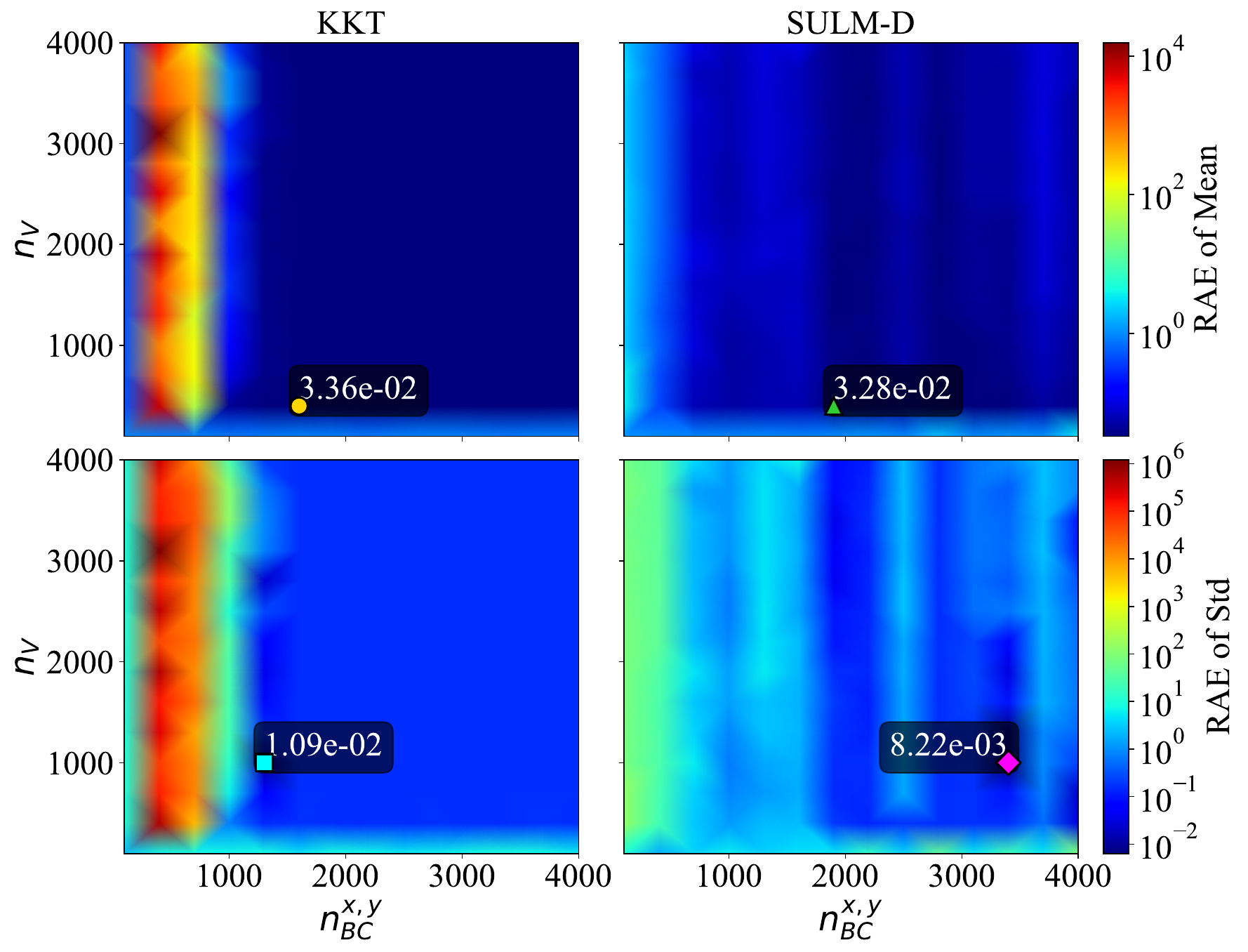}
        \caption{}
        \label{fig:EBKL_heatmap}
    \end{subfigure}
    \begin{subfigure}{0.45\textwidth}
        \centering
        \includegraphics[width=\linewidth]{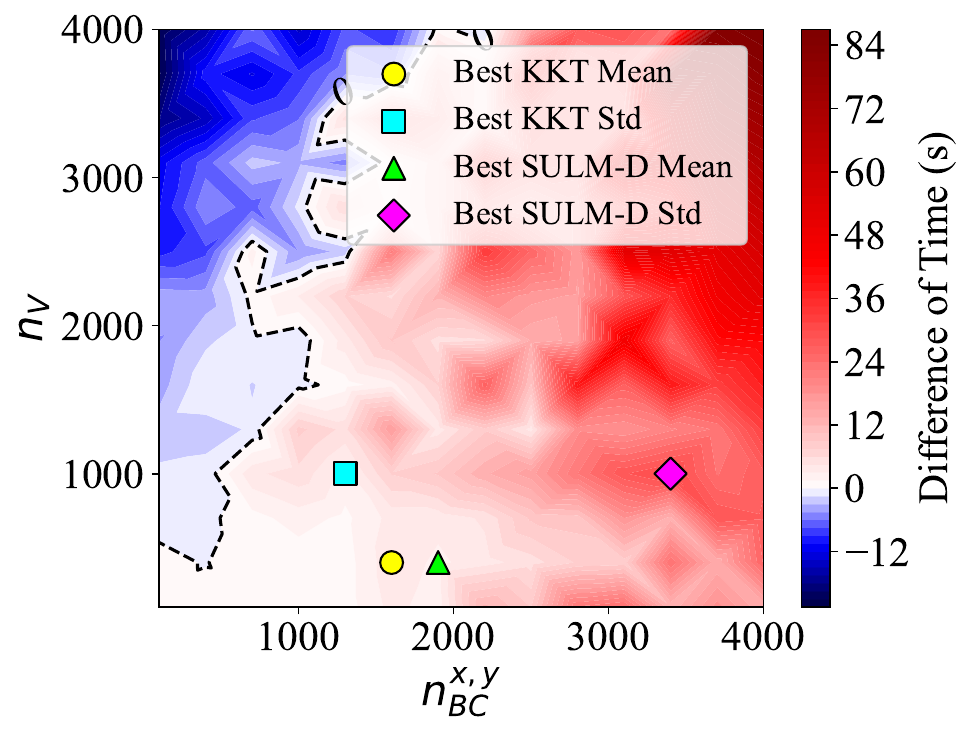}
        \caption{}
        \label{fig:EBKL_diff}
    \end{subfigure}
    \caption{Comparison of (a) computational cost, (b) RAE of mean and standard deviation, and (c) locations of optimal $PC^2$ parameter settings for the 1D Euler equation with Karhunen–Loève expanded inputs, obtained using the KKT and SULM-D methods.}
\end{figure}

For PC$^2$, we use a polynomial order of $p=10$ with a hyperbolic truncation parameter $q=0.7$, as the contribution of three-variable interaction terms is negligible in this example. We investigate $n_{\mathrm{V}}$ and $n{^{x,y}_{BC}}$ ($n_{\mathrm{BC}}$ for each spatial dimension) ranging from 100 to 4000, and no additional data are introduced for training. Fig. \ref{fig:EBKL_two} illustrates the computational cost of KKT and SULM-D under different combinations of $n^{x,y}_{BC}$ and $n_{\mathrm{V}}$. Since the main purpose of \PC is UQ, Fig. \ref{fig:EBKL_heatmap} presents the RAE of the mean and standard deviation of the fitting solutions, which are obtained via MC on the PCE model. Based on these figures, the locations corresponding to the optimal values of these four indicators are identified and subsequently marked in Fig. \ref{fig:EBKL_diff}, which shows the difference in computational cost between the two strategies. It can be observed that, in this example, all optimal points are found within the region where the computational cost of SULM-D is cheaper, which indicates that the cooperation of SULM with D-optimal sampling in higher dimensional problems can leverage the time-saving advantage of SULM to offset the additional computational burden introduced by D-optimal sampling, thereby achieving superior accuracy and efficiency than KKT at the same time.

\subsection{2D Heat Equation with Dirichlet Boundary Conditions and KL-Decomposed Stochastic Source}
\label{subsec35}
In the last example, a 2D heat equation with a stochastic source term given by a spatially dependent random field is discussed as follows:
\vspace{0.6em}
\begin{equation}
\begin{aligned}
    \frac{\partial u(x,y, t)}{\partial t}
    &-\mathcal{D} \left( \frac{\partial^2u(x,y,t)}{\partial x^2}
    +\frac{\partial^2u(x,y,t)}{\partial y^2} \right)
    = f(x,y), 
    \qquad x,y \in [0, 1], \; t \in [0, 1], \\
    u(0,y,t) &= u(1,y,t)= u(x,0,t)= u(x,1,t)=0, \quad u(x,0) = \sin (2\pi x)\sin(2\pi y),
\end{aligned}
\label{Eq:2D_HEKL}
\end{equation}
where thermal diffusivity coefficient is fixed at $\mathcal{D}=0.01$, and $f(x,y)$ is the source term sampled from a zero-mean 2D Gaussian random field (GRF) with radial basis function (RBF) kernel, defined by a standard deviation $\sigma = 0.05$ and a correlation length $l_{c}=0.2$. The GRF is expanded through a KL expansion, where it has been examined that the first 28 modes can capture more than 99\% cumulative variance. However, as the curse of dimensionality leads to an exponential-like increase in computational cost, which is crucial bottleneck especially for KKT solver, and since this example is primarily intended to compare the performance of the two solvers and the D-optimal sampling with increasing dimension, only the first 8 eigenmodes are considered in the following discussion. Therefore, the total number of input variables is 11 in this example.

\begin{figure}[htbp]
    \centering
    \begin{subfigure}{0.49\textwidth}
        \centering        \includegraphics[width=\linewidth]{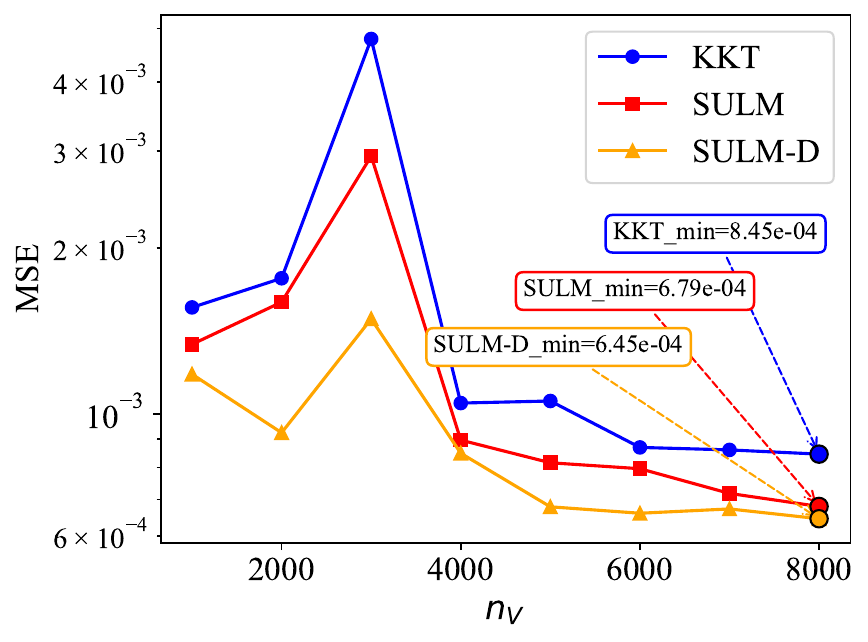}
        \caption{}
        \label{fig:HEKL_MSE}
    \end{subfigure}
    \begin{subfigure}{0.49\textwidth}
        \centering
        \includegraphics[width=\linewidth]{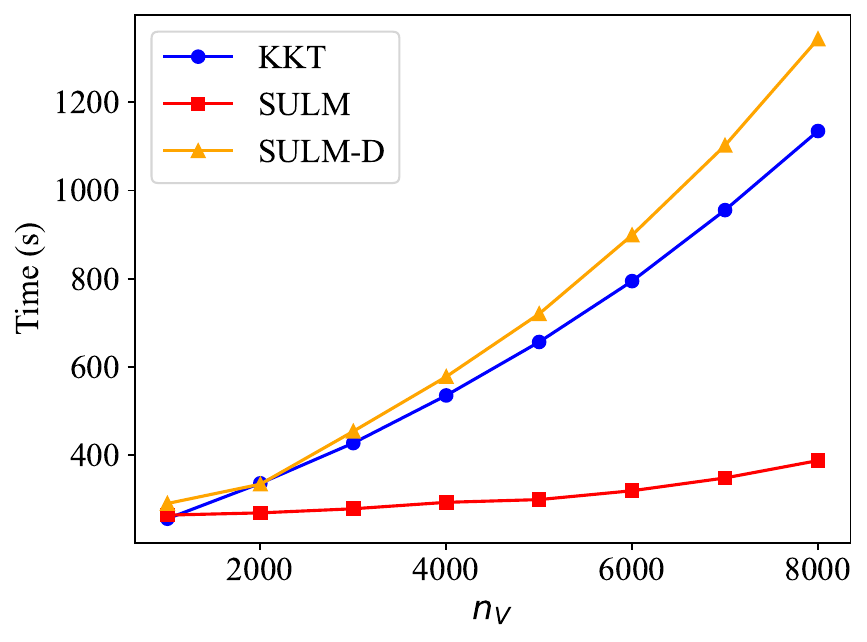}
        \caption{}
        \label{fig:Time_HEKL}
    \end{subfigure}
    \caption{Comparison of (a) computational accuracy and (b) computational cost for the 2D heat equation with Karhunen–Loève expanded random fields of source terms, using the KKT, SULM, and SULM-D solvers.}
\end{figure}

\begin{figure}[t!]
    \centering
    \includegraphics[width=0.6\textwidth]{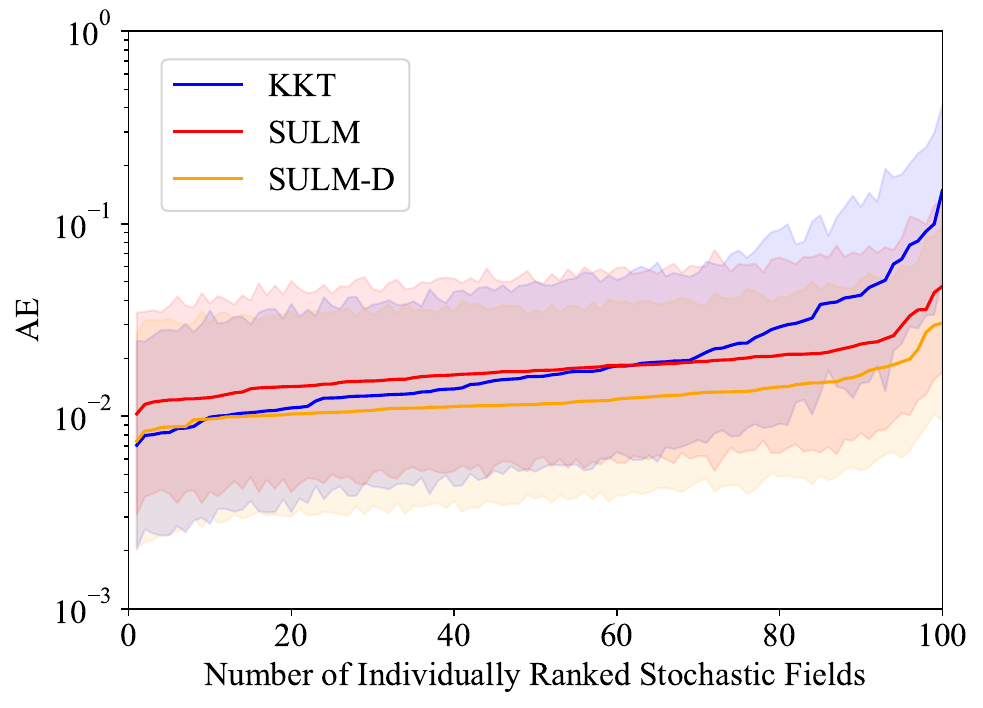}
    \caption{Comparison of individually sorted AE, including mean values and their standard deviation bands, for KKT, SULM, and SULM-D over 100 stochastic fields at $n_{\mathrm{V}}=8000$.}
    \label{fig:HEKL_100RF}
\end{figure}

\begin{figure}[p]
    \centering
    \includegraphics[width=1\textwidth]{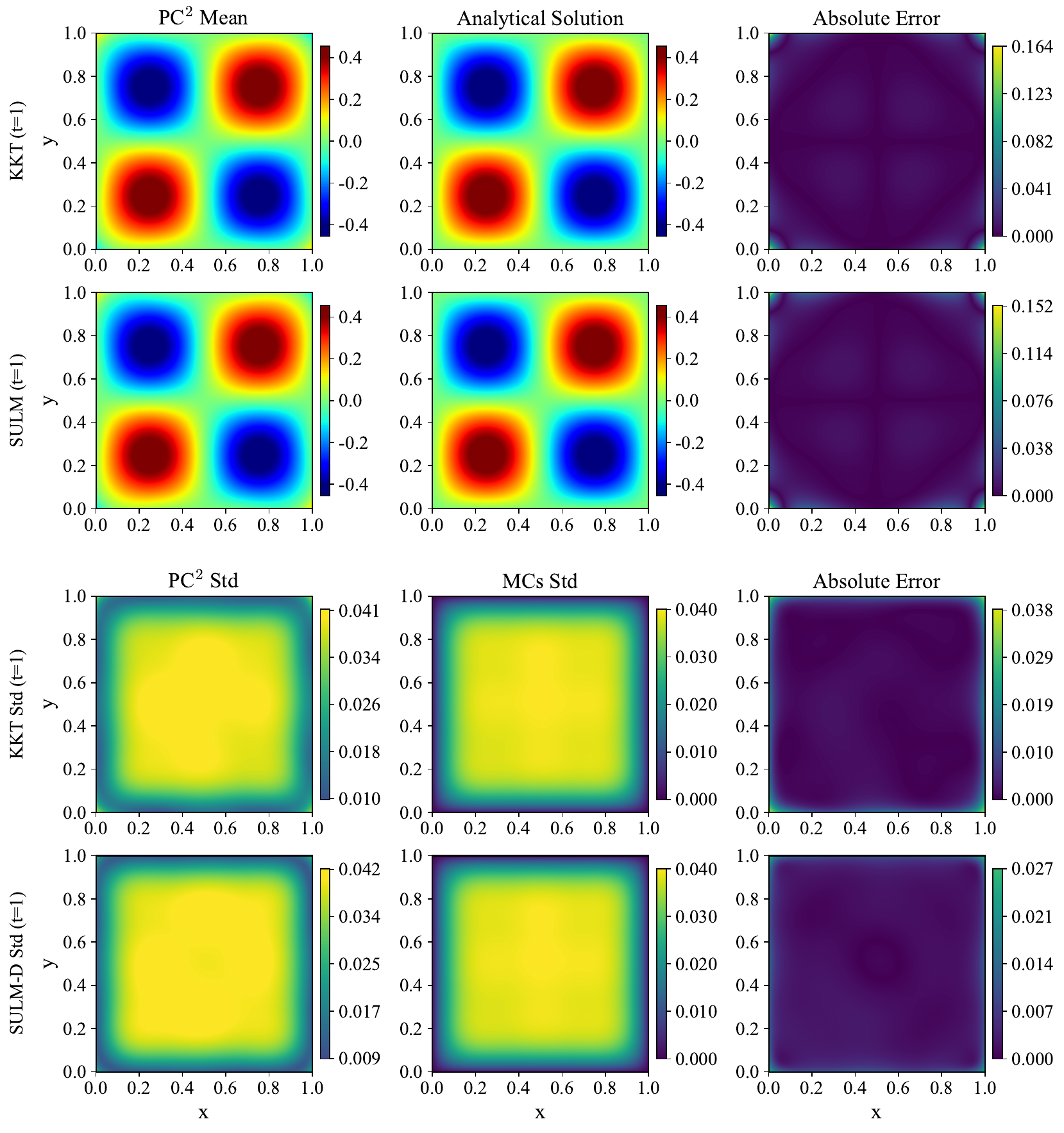}
    \caption{Comparison of the mean and standard deviation of the temperature field at $t=1$ obtained by KKT and SULM with $n_V=8000$. The first and second rows show the mean fields solved by KKT and SULM compared with the analytical solution and their corresponding absolute errors. The third and forth rows show the standard deviation fields solved by KKT and SULM, in comparison with the MC simulation result, together with their absolute error distributions. Note that, for a consistent comparison, all field plots in the third and forth rows share the same colorbar mapping.}
    \label{fig:HEKL_field_comparison}
\end{figure}

For PC$^2$, we use a polynomial order of $p=14$ with a hyperbolic truncation parameter $q=0.6$. We set $n_{\mathrm{BC}}=4000$ (2000 per spatial dimension) and $n_{init}=4000$, while gradually increasing $n_{\mathrm{V}}$ from 1000 to 8000. In this example, it is challenging to achieve satisfactory accuracy using a purely physics-informed strategy. Therefore, a hybrid strategy that combines data-driven and physics-informed components is adopted. Since the original idea of $PC^2$ is to preserve the physics-informed nature rather than relying heavily on data, the full-field numerical results are not directly used for training. Instead, only several grid-node values are randomly selected from these full-field solutions to form the training ($N=5000$) and testing datasets ($N_{}=500$). Both datasets are produced by fine-grid FD simulations, with an expected MSE on the order of $10^{-6}$–$10^{-8}$ arising from spatial and temporal discretization. As shown in Fig. \ref{fig:HEKL_MSE}, the MSE of KKT, SULM, and SULM-D exhibits an initial rise followed by a gradual decline and eventual convergence. The overall accuracy ranking (SULM-D > SULM > KKT) remains consistent with the previous examples, as also reflected by the minimum MSE highlighted in the figure ($8.45\times10^{-4}$ for KKT, $6.79\times10^{-4}$ for SULM, and $6.45\times10^{-4}$ for SULM-D). For the computation cost, at the beginning, the runtimes of all three strategies are nearly identical. As $n_{\mathrm{V}}$ increases, the SULM shows a significantly higher efficiency compared to KKT and SULM-D. In contrast, the overall computational costs of SULM-D and KKT remain at a similar level. To further evaluate and compare the generalization capability of these solves across different stochastic realizations, we generate 100 random source fields through KL expansion and test the trained surrogate models on this set. For each stochastic field, the MSE is computed over a uniform grid of space and time, and their mean values are individually sorted in a ascending order to facilitate a more direct and clear comparison. As shown in Fig.~\ref{fig:HEKL_100RF}, SULM-D achieves the lowest error overall, followed by SULM, while KKT exhibits larger magnitudes and variability of MSE across different random fields. For visualization, Fig.~\ref{fig:HEKL_field_comparison} shows the temperature mean and standard deviation fields at t=1 solved by KKT and SULM using ReducedPCE \cite{NOVAK2024112926}. The mean fields are compared with the analytical deterministic solution, and the corresponding absolute errors are calculated to verify the accuracy of the prediction. The temperature mean fields of KKT and SULM are highly consistent with the analytical solution, with relatively larger errors appearing only near the corners of the boundary. To further evaluate the performance of UQ, a reference standard deviation field is established by combining the FD method with MC simulations, and the result is compared with the standard deviation fields obtained from KKT and SULM. As shown in the third and fourth rows of Fig.~\ref{fig:HEKL_field_comparison}, the results of KKT and SULM exhibit strong consistency with the reference field, simultaneously reflecting the characteristics of  the distribution of the truncated GRF. Only slight deviations are observed at the corners and within extremely narrow boundary regions. Overall, for this example, considering both computational efficiency and accuracy, the SULM represents the most balanced and effective choice. If higher accuracy is desired, the SULM-D can also serve as a suitable alternative to KKT, achieving better precision with only a marginal increase in computational time.

\section{Conclusion}
\label{sec4}
In this work, we develop the physics-informed polynomial chaos expansion (\PC) framework by introducing a numerically efficient constraint optimization solver (SULM) and a D-optimal sampling strategy. The proposed methods maintain the physical interpretability of the standard PC$^2$ while significantly improve computational efficiency and robustness especially in high-dimensional problems. The numerical results of representative physical ordinary and partial differential problems indicate that, for complex or high-dimensional cases, the combination of SULM with D-optimal sampling can achieve comparable or higher accuracy and improved numerical stability relative to the standard PC$^2$, while requiring similar or even less computational time. In addition, by solely replacing the standard PC$^2$ with SULM, the computational cost can be significantly reduced while also maintaining a satisfactory level of accuracy for most cases. Furthermore, with the number of virtual points increasing, SULM consistently converges and effectively avoid the homogenization issue observed in the standard PC$^2$ under certain overconstrained scenarios. 

This study also opens the door to various modifications of the original \PC\ framework to fulfill the specific requirements of the investigated phenomena, such as dimensionality or computational cost. In this work, we demonstrate that a simple replacement of the KKT formulation with SULM leads to numerically efficient solutions, while D-optimal statistical sampling ensures stable convergence as the number of samples increases. Nevertheless, many other potential adaptations remain to be explored, including alternative solvers, sampling strategies, or sparse formulations \cite{LuthenReview}, which will be the subject of our future research.

Overall, the enhanced PC$^2$ framework represents a versatile and efficient approach to physics-informed surrogate modeling. Its flexible structure allows for numerous modifications inspired by existing literature, such as the integration of specific solvers or advanced sampling strategies, thereby enabling adaptation to a wide range of applications.

\section{Acknowledgements}
This work was supported by the Czech Science Foundation under project No. 24-10892S. The international collaboration was partly supported by the Ministry of Education, Youth and Sports of the Czech Republic under project No. LUAUS24260. HS and MDS gratefully acknowledge the support of the Defense Threat Reduction Agency, Award HDTRA12020001.

\bibliographystyle{elsarticle-num}
\bibliography{literatura}

@Article{Alizadeh2020,
  author  = {R. Alizadeh and J. K. Allen and F. Mistree},
  title   = {Managing Computational Complexity Using Surrogate Models: A Critical Review},
  journal = {Research in Engineering Design},
  year    = {2020},
  volume  = {31},
  pages   = {275-298},
  doi     = {10.1007/s00163-020-00336-7},
}

@article{karniadakis2021physicsinformed,
  author = {Karniadakis, George Em and Kevrekidis, Ioannis G. and Lu, Lu and Perdikaris, Paris and Wang, Sifan and Yang, Liu},
  doi = {10.1038/s42254-021-00314-5},
  issn = {25225820},
  journal = {Nature Reviews Physics},
  number = 6,
  pages = {422--440},
  title = {Physics-informed machine learning},
  volume = 3,
  year = 2021
}

@article{RAISSI2019686,
title = {Physics-informed neural networks: A deep learning framework for solving forward and inverse problems involving nonlinear partial differential equations},
journal = {Journal of Computational Physics},
volume = {378},
pages = {686-707},
year = {2019},
issn = {0021-9991},
doi = {https://doi.org/10.1016/j.jcp.2018.10.045},
author = {M. Raissi and P. Perdikaris and G.E. Karniadakis},
}

@article{RAISSI2017683,
title = {Machine learning of linear differential equations using Gaussian processes},
journal = {Journal of Computational Physics},
volume = {348},
pages = {683-693},
year = {2017},
issn = {0021-9991},
doi = {https://doi.org/10.1016/j.jcp.2017.07.050},
author = {Maziar Raissi and Paris Perdikaris and George Em Karniadakis},
}

@article{NOVAK2024112926,
title = {Physics-informed polynomial chaos expansions},
journal = {Journal of Computational Physics},
volume = {506},
pages = {112926},
year = {2024},
issn = {0021-9991},
doi = {https://doi.org/10.1016/j.jcp.2024.112926},
author = {Lukáš Novák and Himanshu Sharma and Michael D. Shields},
}

@article{SHARMA2024117314,
title = {Physics-constrained polynomial chaos expansion for scientific machine learning and uncertainty quantification},
journal = {Computer Methods in Applied Mechanics and Engineering},
volume = {431},
pages = {117314},
year = {2024},
issn = {0045-7825},
doi = {https://doi.org/10.1016/j.cma.2024.117314},
author = {Himanshu Sharma and Lukáš Novák and Michael Shields},
}

@article{Cai2021,
  title={Physics-informed neural networks (PINNs) for fluid mechanics: A review},
  author={Cai, Shengze and Mao, Zhiping and Wang, Zhicheng and Yin, Minglang and Karniadakis, George Em},
  journal={Acta Mechanica Sinica},
  volume={37},
  number={12},
  pages={1727-1738},
  year={2021},
  publisher={Springer}
}

@article{Pang2019,
author = {Pang, Guofei and Lu, Lu and Karniadakis, George Em},
title = {fPINNs: Fractional Physics-Informed Neural Networks},
journal = {SIAM Journal on Scientific Computing},
volume = {41},
number = {4},
pages = {A2603-A2626},
year = {2019},
doi = {10.1137/18M1229845},
}

@INPROCEEDINGS{Misyris9282004,
  author={Misyris, George S. and Venzke, Andreas and Chatzivasileiadis, Spyros},
  booktitle={2020 IEEE Power\& Energy Society General Meeting (PESGM)}, 
  title={Physics-Informed Neural Networks for Power Systems}, 
  year={2020},
  volume={},
  number={},
  pages={1-5},
  doi={10.1109/PESGM41954.2020.9282004}
}

@article{TARTAKOVSKY2023967,
title = {Physics-informed Gaussian process regression for states estimation and forecasting in power grids},
journal = {International Journal of Forecasting},
volume = {39},
number = {2},
pages = {967-980},
year = {2023},
issn = {0169-2070},
doi = {https://doi.org/10.1016/j.ijforecast.2022.03.007},
author = {Alexandre M. Tartakovsky and Tong Ma and David A. Barajas-Solano and Ramakrishna Tipireddy},
}

@article{SCOTT2022,
title = {Solving large linear least squares problems with linear equality constraints},
journal = {BIT Numerical Mathematics},
volume = {62},
pages = {1765–1787},
year = {2022},
issn = {1572-9125},
author = {Jennifer Scott and Miroslav Tůma}
}

@article{TSAPETIS2023101561,
title = {UQpy v4.1: Uncertainty quantification with Python},
journal = {SoftwareX},
volume = {24},
pages = {101561},
year = {2023},
issn = {2352-7110},
author = {Dimitrios Tsapetis and Michael D. Shields and Dimitris G. Giovanis and Audrey Olivier and Lukas Novak and Promit Chakroborty and Himanshu Sharma and Mohit Chauhan and Katiana Kontolati and Lohit Vandanapu and Dimitrios Loukrezis and Michael Gardner}
}

@article{XIANG202211,
title = {Self-adaptive loss balanced Physics-informed neural networks},
journal = {Neurocomputing},
volume = {496},
pages = {11-34},
year = {2022},
issn = {0925-2312},
doi = {https://doi.org/10.1016/j.neucom.2022.05.015},
author = {Zixue Xiang and Wei Peng and Xu Liu and Wen Yao},
}

@article{SAHLICOSTABAL2024107324,
title = {Δ-PINNs: Physics-informed neural networks on complex geometries},
journal = {Engineering Applications of Artificial Intelligence},
volume = {127},
pages = {107324},
year = {2024},
issn = {0952-1976},
doi = {https://doi.org/10.1016/j.engappai.2023.107324},
author = {Francisco {Sahli Costabal} and Simone Pezzuto and Paris Perdikaris},
}

@article{moseley2023finite,
  title={Finite basis physics-informed neural networks (FBPINNs): a scalable domain decomposition approach for solving differential equations},
  author={Moseley, Ben and Markham, Andrew and Nissen-Meyer, Tarje},
  journal={Advances in Computational Mathematics},
  volume={49},
  number={4},
  pages={62},
  year={2023},
  publisher={Springer}
}

@article{li2022gradient,
  title={Gradient-optimized physics-informed neural networks (GOPINNs): a deep learning method for solving the complex modified KdV equation},
  author={Li, Jiaheng and Chen, Junchao and Li, Biao},
  journal={Nonlinear Dynamics},
  volume={107},
  number={1},
  pages={781-792},
  year={2022},
  publisher={Springer}
}

@article{LuthenReview,
author = {L\"{u}then, Nora and Marelli, Stefano and Sudret, Bruno},
title = {Sparse Polynomial Chaos Expansions: Literature Survey and Benchmark},
journal = {SIAM/ASA Journal on Uncertainty Quantification},
volume = {9},
number = {2},
pages = {593-649},
year = {2021},
doi = {10.1137/20M1315774}
}

@article{OLADYSHKIN2012179,
title = {Data-driven uncertainty quantification using the arbitrary polynomial chaos expansion},
journal = {Reliability Engineering \& System Safety},
volume = {106},
pages = {179-190},
year = {2012},
issn = {0951-8320},
doi = {https://doi.org/10.1016/j.ress.2012.05.002},
author = {S. Oladyshkin and W. Nowak}
}

@article{BLATMAN20112345,
title = {Adaptive sparse polynomial chaos expansion based on least angle regression},
journal = {Journal of Computational Physics},
volume = {230},
number = {6},
pages = {2345-2367},
year = {2011},
issn = {0021-9991},
doi = {https://doi.org/10.1016/j.jcp.2010.12.021},
author = {Géraud Blatman and Bruno Sudret}
}

@misc{gaynutdinova2025homogenizationguaranteedboundsprimaldual,
      title={Homogenization with Guaranteed Bounds via Primal-Dual Physically Informed Neural Networks}, 
      author={Liya Gaynutdinova and Martin Doškář and Ondřej Rokoš and Ivana Pultarová},
      year={2025},
      eprint={2509.07579},
      archivePrefix={arXiv},
      primaryClass={cs.LG},
      url={https://arxiv.org/abs/2509.07579}, 
}

@article{JAKEMANDependentPCE,
title = {Polynomial chaos expansions for dependent random variables},
journal = {Computer Methods in Applied Mechanics and Engineering},
volume = {351},
pages = {643-666},
year = {2019},
issn = {0045-7825},
doi = {https://doi.org/10.1016/j.cma.2019.03.049},
author = {John D. Jakeman and Fabian Franzelin and Akil Narayan and Michael Eldred and Dirk Plfüger}
}

@Article{Askey,
	title = {The {W}iener--{A}skey Polynomial Chaos for Stochastic Differential Equations},
	year = {2002},
  doi = {10.1137/s1064827501387826},
  volume = {24},
  number = {2},
  pages = {619--644},
  author = {Dongbin Xiu and George Em Karniadakis},
  journal = {{SIAM} Journal on Scientific Computing},
}

@article{nabian2021efficient,
  title={Efficient training of physics-informed neural networks via importance sampling},
  author={Nabian, Mohammad Amin and Gladstone, Rini Jasmine and Meidani, Hadi},
  journal={Computer-Aided Civil and Infrastructure Engineering},
  volume={36},
  number={8},
  pages={962-977},
  year={2021},
  doi = {https://doi.org/10.1111/mice.12685}
}

@article{guo2022novel,
  title={A novel adaptive causal sampling method for physics-informed neural networks},
  author={Guo, Jia and Wang, Haifeng and Hou, Chenping},
  journal={arXiv preprint arXiv:2210.12914},
  year={2022}
}

@article{sharma2025polynomial,
  title={Polynomial chaos expansion for operator learning},
  author={Sharma, Himanshu and Nov{\'a}k, Luk{\'a}{\v{s}} and Shields, Michael D},
  journal={arXiv preprint arXiv:2508.20886},
  year={2025},
  doi = {https://doi.org/10.48550/arXiv.2508.20886}
}

@book{ghanem2003stochastic,
  title={Stochastic finite elements: a spectral approach},
  author={Ghanem, Roger G and Spanos, Pol D},
  year={2003},
  publisher={Courier Corporation}
}

@article{loeppky2009choosing,
  title={Choosing the sample size of a computer experiment: A practical guide},
  author={Loeppky, Jason L and Sacks, Jerome and Welch, William J},
  journal={Technometrics},
  volume={51},
  number={4},
  pages={366--376},
  year={2009},
  publisher={Taylor \& Francis}
}

@Inbook{Goswami2023,
author="Goswami, Somdatta
and Bora, Aniruddha
and Yu, Yue
and Karniadakis, George Em",
editor="Rabczuk, Timon
and Bathe, Klaus-J{\"u}rgen",
title="Physics-Informed Deep Neural Operator Networks",
bookTitle="Machine Learning in Modeling and Simulation: Methods and Applications",
year="2023",
publisher="Springer International Publishing",
address="Cham",
pages="219--254",
isbn="978-3-031-36644-4",
doi="10.1007/978-3-031-36644-4_6"
}

@article{WU2023115671,
title = {A comprehensive study of non-adaptive and residual-based adaptive sampling for physics-informed neural networks},
journal = {Computer Methods in Applied Mechanics and Engineering},
volume = {403},
pages = {115671},
year = {2023},
issn = {0045-7825},
doi = {https://doi.org/10.1016/j.cma.2022.115671},
author = {Chenxi Wu and Min Zhu and Qinyang Tan and Yadhu Kartha and Lu Lu},
}

@article{vovrechovsky2020modification,
  title={Modification of the maximin and $\phi$ p (phi) criteria to achieve statistically uniform distribution of sampling points},
  author={Vo{\v{r}}echovsk{\`y}, Miroslav and Eli{\'a}{\v{s}}, Jan},
  journal={Technometrics},
  volume={62},
  number={3},
  pages={371--386},
  year={2020},
  publisher={Taylor \& Francis}
}

@book{fang1993number,
  title={Number-theoretic methods in statistics},
  author={Fang, Kai-Tai and Wang, Yuan},
  volume={51},
  year={1993},
  publisher={CRC Press}
}

@incollection{hickernell1998lattice,
  title={Lattice rules: how well do they measure up?},
  author={Hickernell, Fred J},
  booktitle={Random and quasi-random point sets},
  pages={109--166},
  year={1998},
  publisher={Springer}
}

@article{McKay01051979,
author = {M. D. McKay and R. J. Beckman and W. J. Conover},
title = {Comparison of Three Methods for Selecting Values of Input Variables in the Analysis of Output from a Computer Code},
journal = {Technometrics},
volume = {21},
number = {2},
pages = {239--245},
year = {1979},
publisher = {ASA Website},
doi = {10.1080/00401706.1979.10489755},
}

@article{johnson1990minimax,
  title={Minimax and maximin distance designs},
  author={Johnson, Mark E and Moore, Leslie M and Ylvisaker, Donald},
  journal={Journal of statistical planning and inference},
  volume={26},
  number={2},
  pages={131--148},
  year={1990},
  publisher={Elsevier}
}

@article{morris1995exploratory,
  title={Exploratory designs for computational experiments},
  author={Morris, Max D and Mitchell, Toby J},
  journal={Journal of statistical planning and inference},
  volume={43},
  number={3},
  pages={381--402},
  year={1995},
  publisher={Elsevier}
}

\end{document}